\documentclass[10pt,twocolumn,letterpaper]{article} 
\usepackage{cvpr}
\usepackage{times}
\usepackage{epsfig}
\usepackage{graphicx}
\usepackage{amsmath}
\usepackage{amssymb}
\usepackage{gensymb}
\usepackage[table,xcdraw]{xcolor}
\usepackage{cite}
\usepackage{footmisc}
\usepackage{enumitem}
\usepackage{multirow}
\usepackage{nth}
\usepackage{adjustbox}

\usepackage[pagebackref=true,breaklinks=true,letterpaper=true,colorlinks,bookmarks=false]{hyperref}

\cvprfinalcopy 

\def\cvprPaperID{5194} 

\ifcvprfinal\fi
\begin{document}

\title{SpherePHD: Applying CNNs on a Spherical PolyHeDron Representation of 360\degree{} Images} 

\author{Yeonkun Lee\thanks{These authors contributed equally} ,   Jaeseok Jeong\footnotemark[1] ,   Jongseob Yun\footnotemark[1] ,   Wonjune Cho\footnotemark[1] , Kuk-Jin Yoon\\
Visual Intelligence Laboratory, Department of Mechanical Engineering, KAIST, Korea\\
\tt\small \{dldusrjs, jason.jeong, jseob, wonjune, kjyoon\}@kaist.ac.kr
}

\maketitle
\thispagestyle{empty}

\begin{abstract}
    Omni-directional cameras have many advantages over conventional cameras in that they have a much wider field-of-view (FOV). Accordingly, several approaches have been proposed recently to apply convolutional neural networks (CNNs) to omni-directional images for various visual tasks. However, most of them use image representations defined in the Euclidean space after transforming the omni-directional views originally formed in the non-Euclidean space. This transformation leads to shape distortion due to nonuniform spatial resolving power and the loss of continuity. These effects make existing convolution kernels experience difficulties in extracting meaningful information. 
    
    This paper presents a novel method to resolve such problems of applying CNNs to omni-directional images. The proposed method utilizes a spherical polyhedron to represent omni-directional views. This method minimizes the variance of the spatial resolving power on the sphere surface, and includes new convolution and pooling methods for the proposed representation. The proposed method can also be adopted by any  existing CNN-based methods. The feasibility of the proposed method is demonstrated through classification, detection, and semantic segmentation tasks with synthetic and real datasets. 

\end{abstract}

\section{Introduction} \label{sec:Introduction}

360\degree{} cameras have many advantages over traditional cameras because they offer an omni-directional view of a scene rather than a narrow field of view. This omni-directional view of 360\degree{} cameras\footnote{`360\degree{}' and `omni-directional' are used interchangeably in the paper.} allows us to extract more information from the scene at once. Therefore, 360\degree{} cameras play an important role in systems requiring rich information of surroundings, \eg advanced driver assistance systems (ADAS) and autonomous robotics. 

Meanwhile, convolutional neural networks (CNNs) have been widely used for many visual tasks to preserve locality information. They have shown great performance in classification, detection, and semantic segmentation problems as in  \cite{NIPS2012_4824}\cite{DBLP:journals/corr/LongSD14}\cite{Maturana-2015-6018}\cite{Redmon_2016_CVPR}\cite{DBLP:journals/corr/RenHG015}. 

Following this trend, several approaches have been proposed recently to apply CNNs to omni-directional images to solve  classification, detection, and semantic segmentation problems. Since the input data of neural networks are usually represented in the Euclidean space, they need to represent the omni-directional images in the Euclidean space, even though omni-directional images are originally represented in non-Euclidean polar coordinates. 
Despite omni-directional images existing in a non-Euclidean space, equi-rectangular projection (ERP) has been commonly used to represent omni-directional images in regular grids. 

\begin{figure}[t]
    \begin{centering}
      \includegraphics[width=1.0\linewidth]{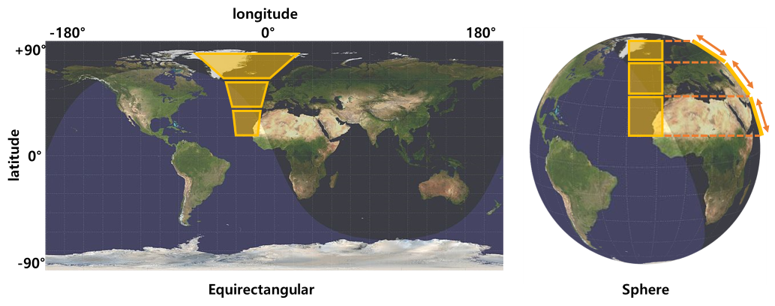}
      \end{centering}
      \vspace{-10pt}
      \caption{Spatial distortion due to  nonuniform spatial resolving power in an ERP image. Yellow squares on both sides represent the same surface areas on the sphere.}
    \vspace{-6pt}
    \label{fig:earth}
\end{figure}

\begin{figure}[t]
    \begin{center}
      \includegraphics[width=0.49\linewidth]{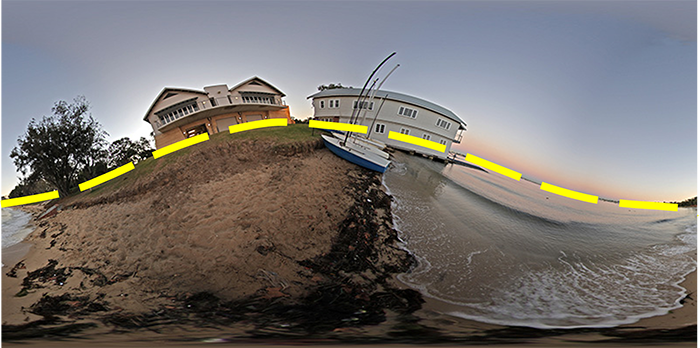}
        \includegraphics[width=0.49\linewidth]{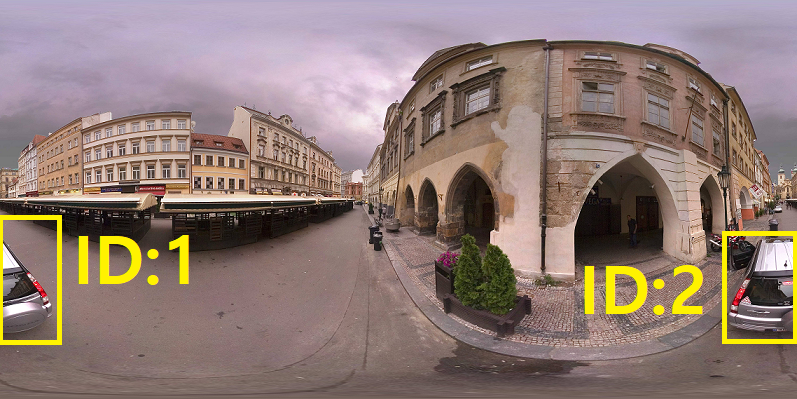}
    \end{center}
    \vspace{-6pt}
      \caption{Problems of using ERP images. (left) When an omni-directional image is taken by a tilted camera, a sinusoidal fluctuation of the horizon occurs in the ERP image. The yellow dotted line represents the fluctuating horizon\cite{Sphere2Sphere}. (right) When the car is split, the car is detected as cars with two different IDs. This image, which is from the SUN360 dataset\cite{SUN360}, demonstrate the effects of edge discontinuity. 
      }
    \label{fig:tilted}
    \vspace{-6pt}
\end{figure}

However, there exists spatial distortion in ERP images coming from the nonuniform spatial resolving power, which is caused by the non-linearity of the transformation as shown in Fig.~\ref{fig:earth}. This effect is more severe near the poles. In addition, a different kind of distortion is observable when the omni-directional image is taken by a tilted 360\degree{} camera. In this case, without compensating for the camera tilt angle, ERP introduces a sinusoidal fluctuation of the horizon as in Fig.~\ref{fig:tilted}. This also distorts the image, making the task of visual perception much more difficult.

In conventional 2D images, the regular grids along vertical and horizontal directions allow the convolution domain to be uniform; the uniform domain enables the same shaped convolution kernels to be applied over the whole image. However, in ERP images, the nonuniform spatial resolving power, as shown in Fig.~\ref{fig:earth} and  Fig.~\ref{fig:tilted}(left), causes the convolution domain to vary over the ERP image, making the same shaped kernels inadequate for the ERP image convolution. 

Furthermore, during the transformation from the non-Euclidean space to the Euclidean space, some important properties could be lost. For example, the non-Euclidean space in which omni-directional images are formed has a cyclical property: the unidirectional translation along a sphere surface (\ie in an omni-directional image) is always continuous and returns to the starting point eventually. However, during the transformation from the non-Euclidean space to the Euclidean space through the ERP, the continuity and cyclical properties are lost. This causes a discontinuity along the borders of the ERP image. The discontinuity can cause a single object to be misinterpreted in detection problems as two different objects when the object is split by the borders as shown in Fig.~\ref{fig:tilted}(right). 

Recently, there has been an attempt to keep the continuity of the omni-directional image by projecting the image onto a cube map \cite{monroy2018salnet360}. Projecting an omni-directional image onto a cube map also has the benefit that the spatial resolving power in an image varies much less compared with ERP images. Furthermore, the cube map representation is affected less by rotation than the ERP representation. However, even though a cube map representation reduces the variance of spatial resolving power, there still exists variance from centers of the cube faces to their edges. In addition, the ambiguity of kernel orientation exists when a cube map representation is applied to CNNs. Because the top and bottom faces of cube map are orthogonal to the other faces, it is ambiguous to define kernel orientation to extract uniform locality information on top and bottom faces. The effects of these flaws are shown in Sec.\ref{sec:Experiments}.

In this paper, we propose a new representation of 360\degree{} images followed by new convolution and pooling methods to apply CNNs to 360\degree{} images based on the proposed representation. In an attempt to reduce the variance of spatial resolving power in representing 360\degree{} images, we come up with a spherical polyhedron-based representation of images (SpherePHD). Utilizing the properties of the SpherePHD constructed by an icosahedral geodesic polyhedron, we present a geometry on which a 360\degree{} image can be projected. 
The proposed projection results in less variance of the spatial resolving power and distortion than others. Furthermore, the rotational symmetry of the spherical geometry allows the image processing algorithms to be rotation invariant. Lastly, the spherical polyhedron provides a continuity property; there is no border where the image becomes discontinuous. This particular representation aims to resolve the issues found in using ERP and cube map representations. The contributions of this paper also include \emph{designing a convolution kernel} and using a specific method of \emph{applying convolution and pooling kernels} for use in CNNs on the proposed spherical polyhedron representation of images. To demonstrate that the proposed method is superior to ERP and cube map representations, we compare classification, detection, and semantic segmentation accuracies for the different representations. To conduct comparison experiments, we also created \emph{a spherical MNIST dataset}\cite{deng2012mnist}, \emph{spherical SYNTHIA dataset} \cite{Ros_2016_CVPR}, and \emph{spherical Stanford2D3D dataset} \cite{stanford2D3D} through the spherical polyhedron projection of the original datasets. The source codes for our method are available at \url{https://github.com/KAIST-vilab/SpherPHD_public}.

\section{Related works} \label{sec:Related works}
In this section, we discuss relevant studies that have been done for CNNs on omni-directional images.

\subsection{ERP-based methods} 
As mentioned earlier, an ERP image has some flaws: the nonuniform spatial resolving power brings about distortion effects, rotational constraints, and discontinuity at the borders.  
Yang \etal \cite{yang2018object} compared the results of different detection algorithms that take ERP images directly as inputs, showing that not solving the distortion problem still produces relevant accuracy. Other papers proposed to remedy the flaws of ERP-based methods. To tackle the issue of nonuniform resolving power, Coors \etal \cite{Coors_2018_ECCV} proposed sampling pixels from the ERP image, in a rate dependent on latitude, to preserve the uniform spatial resolving power and to keep the convolution domain consistent in each kernel.
Hu \etal \cite{hu2017deep} and Su and Grauman \cite{su2017making} partitioned an omni-directional image into subimages and produced normal field of view (NFoV) images; during the partitioning stages, they also reduced the distortion effects in the ERP image. Lai \etal \cite{lai2018semantic} and Su \etal  \cite{su2016pano2vid} generated saliency maps from omni-directional images and extracted specific NFoV images from areas of high saliency. The use of NFoV image sampling increased accuracy by solving the distortion problem, but extracting NFoV images from an omni-directional image is not as beneficial as utilizing the whole omni-directional view for visual processing. In an attempt to use the whole omni-directional image and also resolve the distortion problem, Su and Grauman \cite{su2017learning} and Tateno \etal \cite{Tateno_2018_ECCV} used distortion-aware kernels at each latitude of the omni-directional image to take into account the resolving power variation. To tackle the distortion problem from tilting the camera, Bourke \cite{Sphere2Sphere} proposed a way to post-process the ERP image, but this requires knowledge of the camera orientation. 
Our method replaces the ERP-based representation as a way to tackle the distortion problems raised from ERP images.

\begin{figure}[t]
    \begin{center}
      \includegraphics[width=0.99\linewidth]{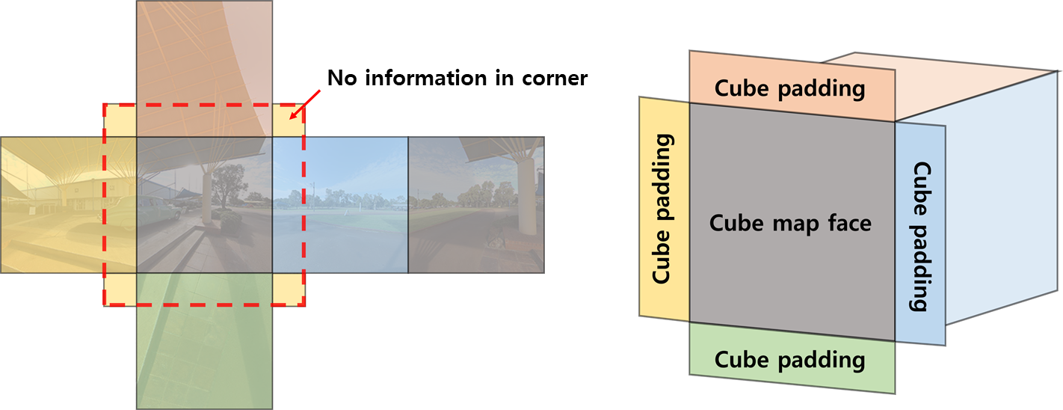}
    \end{center}
    \vspace{-2pt}
      \caption{Cube padding \cite{Cheng_2018_CVPR}. Cube padding allows the receptive field of each face to extend across the adjacent faces.}
    \vspace{-4pt}
    \label{fig:cubepadding}
\end{figure}

\subsection{Other representations for 360$^\circ$ images} 
As a way to represent spherical geometry, a few methods have been proposed to project an omni-directional image onto a cube to generate a cube map. The method in \cite{monroy2018salnet360} utilized a cube map to generate a saliency map of the omni-directional image. As an extension to \cite{monroy2018salnet360}, Cheng \etal \cite{Cheng_2018_CVPR} proposed padding each face with pixels from tangent faces to consider information from all tangent faces and to convolve across the edges. Figure~\ref{fig:cubepadding} shows an example of such a cube padding method. Utilizing the cube map projection of omni-directional images resolves many of the issues that ERP-based methods face. However, in the cube map representation, there still exists noticeable variance of spatial resolving power between the centers and edges of the cube faces. To address this variance of spatial resolving power, Brown \cite{GoogleEquiAngularCube} proposed the equi-angular cube (EAC) projection, changing the method in which omni-directional views are sampled onto a cube. 
Cohen \etal \cite{cohen2018spherical} suggested transforming the domain space from Euclidean $S^2$ space to a SO(3) 3D rotation group to reduce the negative effects of the ERP representation. Compared with the aforementioned approaches, our method further minimizes the variance of the spatial resolving power, while not having to transform into non-spatial domain.

\subsection{Representations in geography} 
Projection of an omni-directional image is also a well known problem in geographical map projection. Throughout history there have been countless map projection methods proposed. Some of the representations mentioned above are examples of these different map projection methods: the work done by Yang \etal \cite{yang2018object} is similar to the Mercator method, work done by Coors \etal \cite{Coors_2018_ECCV} is similar to the Hammer method. The method we propose in this paper is similar to the Dymaxion map projection method.

\section{Proposed method} \label{sec:Our method}

To obtain a new representation of an omni-directional image, we project an omni-directional image onto an icosahedral spherical polyhedron. After the projection, we apply the transformed image to a CNN structure. The advantage of using our representation over using ERP or any other representations is that ours has far less irregularity on an input image.
    
   \subsection{Definition of irregularity} \label{subsec:distortion}
   
   To discuss the irregularity of omni-directional image representations, we need to define a quantitative measure of irregularity. 
   To do so, we first define an effective area of a pixel for a given representation (\ie, ERP, cube map) as the corresponding area when the pixel is projected onto a unit sphere. The irregularity of the omni-directional image representation can then be measured by the variation of the pixels' effective areas. To compare the irregularity for different representations, we define the average effective area of all the pixels in the given representation to be the geometric mean of them as shown in Eq.~(\ref{eq:mean_area}), where $N$ is the total number of pixels and $A_i$ is the effective area of the $i^{th}$ pixel. 
    \begin{equation}\label{eq:mean_area}
        {{A}_{mean}} = \sqrt[N]{\prod_{i=1}^{N}({{A}_{i}})}
    \end{equation}
   Then, we define the irregularity for the $i^{th}$ pixel, $d_i$, as the log scaled ratio of individual pixel's area to their average, shown in Eq.~(\ref{eq:distortion}). 
    \begin{equation}\label{eq:distortion}
        d_i = \log(\frac{{A}_{i}}{{A}_{mean}})
    \end{equation}
   Having defined the mean area to be the geometric mean, the log scaled ratios of irregularity always sum up to zero ($\sum_{i=1}^N d_i =0$). This is a desired behavior of irregularity values because it is a measure of how much each individual pixel's effective area deviates from the average of  pixels' effective areas. We then define an overall irregularity score for each representation as the root-mean-square of the irregularity values as 
    \begin{equation}\label{eq:overall_distortion}
        \mbox{Irregularity score} = \sqrt{\frac{1}{N}\sum_{i=1}^{N}d_i^2} .
    \end{equation}

\subsection{Spherical polyhedron} \label{subsec:SP}

   Spherical polyhedrons are divisions of a sphere by arcs into bounded regions, and there are many different ways to construct such spherical polyhedrons. The aforementioned cube map representation can be regarded as an example of an omni-directional image projected onto a spherical polyhedron. The cube map representation splits the sphere into 6 equi-faced regions, creating a cubic spherical polyhedron; each of these 6 regions represents the face of a cube. Similarly, we can apply such representation to other regular polyhedrons to create a spherical polyhedron that is divided into more equi-faced regions. 
   
\begin{figure}[t]
    \begin{center}
       \includegraphics[width=1.0\linewidth]{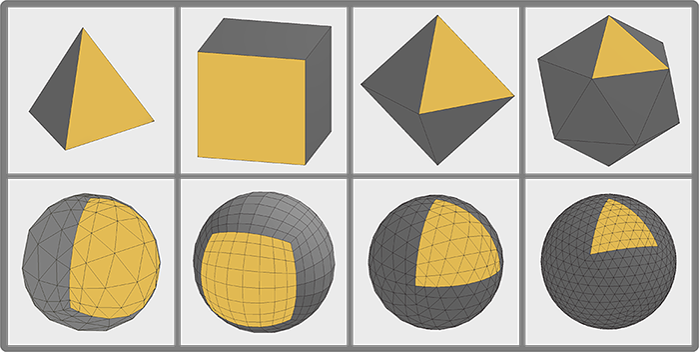}
    \end{center} \vspace{-10pt}
   \caption{Pixel areas on the 3$^{rd}$ subdivision regular convex polyhedrons: tetrahedron, cube, octahedron, and icosahedron. The variance of pixel areas decreases with more equi-faced regions.}
    \vspace{-4pt}
    \label{fig:polyhedron}
    \end{figure}

   In the cube map representation, the squares of a cube are subdivided into smaller squares to represent pixels in a cube map. The irregularity of the cube map representation occurs when we create the cubic spherical polyhedron from a cube map:
   pixels in a cube map correspond to different areas on the cubic spherical polyhedron depending on pixels' locations in the cube map. 
   Here, the irregularity score of the cubic spherical polyhedron converges at an early stage of subdivision\footnote{The irregularity table is included in the supplementary materials\label{footnote_1}} when the squares on the cube are further subdivided into even smaller squares. It means that the irregularity score of a spherical polyhedron is much more dependent on the intrinsic geometry used to make the spherical polyhedron than the number of regular convex polyhedron face subdivisions. This suggests that a regular convex polyhedron with a greater number of faces can have much lower irregularity when it is subdivided and turned into a spherical polyhedron: so we use a regular polyhedron with the most faces, an \emph{icosahedron}. 
   We can visually compare the variances in Fig.~\ref{fig:polyhedron}.

   \subsection{SpherePHD: Icosahedral spherical polyhedron representation} \label{subsec:Icosa}

   When creating a spherical polyhedron using an icosahedron, each subdivision subdivides each triangular face into 4 smaller triangles with the same area. Each of these subdivisions of triangles will be referred to as the $n^{th}$ subdivision where $n$ refers to the number of times the triangles have been subdivided. 
   After creating the subdivided icosahedron, we extrude the newly created vertices from the subdivision onto a sphere, creating a geodesic icosahedron (Note that the original vertices of the icosahedron already lie on a sphere). We can then create a spherical polyhedron through tessellation 
    of this geodesic icosahedron onto a sphere. A spherical polyhedron\footnote{The mentioning of spherical polyhedron throughout this paper will refer specifically to the icosahedral spherical polyhedron (SpherePHD) constructed through a geodesic icosahedron, unless specified otherwise.}
   constructed from a regular convex icosahedron has a smaller irregularity score than the cubical spherical polyhedron\footref{footnote_1}.
   
   Ultimately, we use this SpherePHD to represent omni-directional images. To do so, we take an omni-directional image and project it onto the SpherePHD representation. In this projection, the omni-directional images are represented by the individual triangles tessellated on the SpherePHD. 
   Our SpherePHD representation results in much less variance of effective pixel areas compared with the cube map representation. The minimal variance of effective pixel areas would mean that our representation has less variance in resolving power than the cube representation, which has a lower irregularity score. 
   
   Because the SpherePHD is made from a rotationally symmetric icosahedron, the resulting SpherePHD also has rotationally symmetric properties. 
   In addition, as the projection is a linear transformation, our method does not have any sinusoidal fluctuation effects like those seen in the ERP representation, making our representation more robust to rotation.

   \subsection{SpherePHD convolution and pooling} \label{subsec:SPconv}
   The conventional CNNs compute a feature map through each layer based on a convolution kernel across an image. In order to apply CNNs to SpherePHD, it is necessary to design special convolution and pooling kernels that meet certain criteria as follows: 

   \begin{enumerate}[leftmargin=*]
       \vspace{-6pt}
       \item The convolution kernel should be applicable to all pixels represented in the SpherePHD. 
       \vspace{-4pt}
       \item  The convolution kernel should have the pixel of interest at the center. The output of convolution should maintain the locality information of each pixel and its neighbors without bias.
       \vspace{-4pt}
       \item The convolution kernel should take into account that adjacent triangles are in different orientations.
       \vspace{-4pt}
       \item The pooling kernel should reduce the image from a higher subdivision to a lower subdivision of the SpherePHD.
   \end{enumerate}

   To meet the first condition, our kernel needs to be a collection of triangles as our SpherePHD has triangular pixels. Our kernel must also be applicable to the 12 vertices that are connected to 5 triangles while still being applicable to all the other vertex points connected to 6 triangles. For this reason, the vertices of the kernel should not be connected to more than 5 triangles; if the vertices are connected to more than 5 triangles, those vertices would not be able to convolve at the triangles connected to the 12 original icosahedron's vertices. 

   To satisfy the second condition, the triangular pixel that is being convolved should be at the center of the kernel. If not, the output of the convolution on pixel of interest would be assigned to its neighboring pixel causing a shift for the given pixel. On the spherical surface, when such assignment happens, the direction of shifting varies for all of the convolution output and results in a disoriented convolution output.

   Concerning the third condition, because triangular pixels adjacent to each other in our SpherePHD are oriented differently, we must design a kernel that can be applied to each of the triangles in their respective orientations. 
   
   For the last condition, the pooling kernel can be shaped so that the $n^{th}$ subdivision polyhedron can be downsampled into the $(n-1)^{th}$ subdivision polyhedron. To design such a pooling kernel, we reverse the method in which the subdivisions are formed. 
   In the pooling stage, the pooling kernel is dependent on how the triangles are subdivided in the construction process. For example, if we subdivide a triangle into ${n^2}$ smaller triangles, we can use a pooling kernel that takes the same ${n^2}$ smaller triangles to form one larger triangle. However, for simplicity we subdivide a triangle into 4 (${2^2}$) smaller triangles.
   
   \begin{figure}[t]
    \begin{center}
       \includegraphics[width=0.9\linewidth]{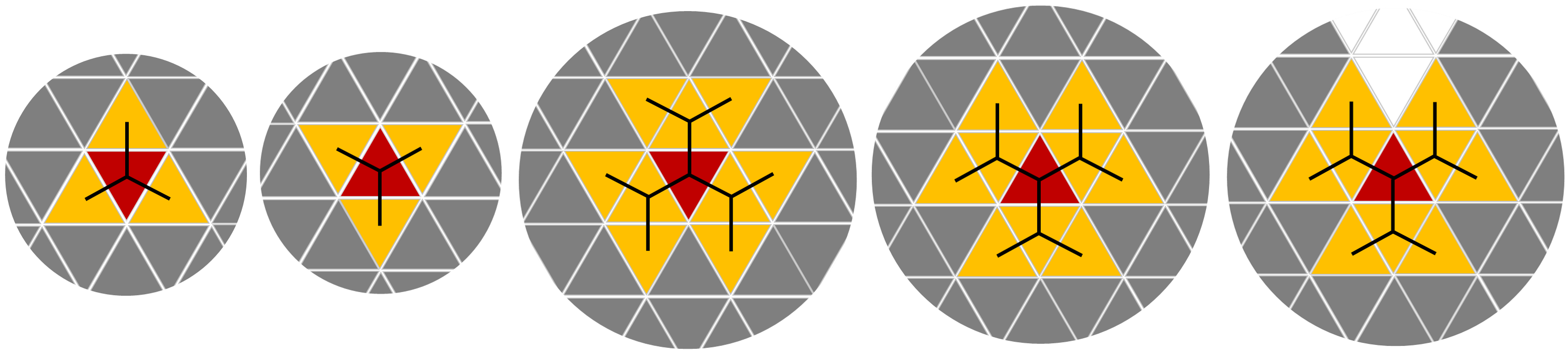}
    \end{center} \vspace{-6pt}
    \caption{From left to right: proposed pooling kernel shape, the same pooling kernel applied onto the adjacent triangle, proposed convolution kernel shape, the same convolution kernel applied onto the adjacent triangle, and the convolution kernel on 12 original vertices of icosahedron (for better understanding, see the yellow kernel in Fig. ~\ref{fig:unfolded}).}\vspace{-2pt}
    \label{fig:kernelshape}
    \end{figure}
    
\begin{figure}[t]
        \begin{center}
        \hspace{0pt}\includegraphics[width=1.0\linewidth]{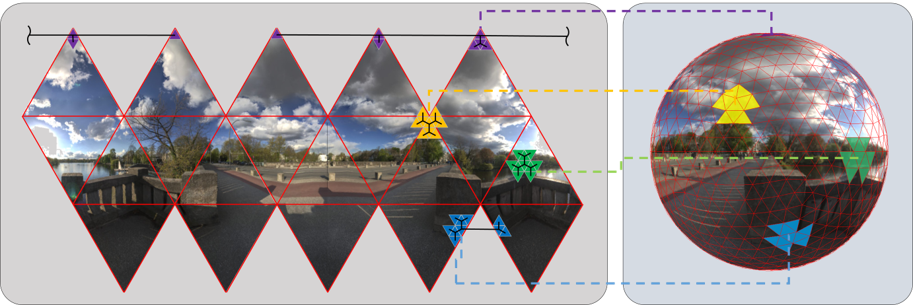}
        \end{center}
        \vspace{-8pt}
        \caption{The visualization of how the kernel is applied to SpherePHD representation. Yellow kernel shows the case when the kernel is located at the vertex of the icosahedron. Purple kernel shows the case when the kernel is located at the pole.}
        \label{fig:unfolded}
    \end{figure}

   Our convolution kernel and pooling kernel that meet the above criteria are given in Fig.~\ref{fig:kernelshape}. These kernels are applied in the same manner to every pixel of the SpherePHD representation. Figure~\ref{fig:unfolded} shows how to apply the proposed kernel to the SpherePHD representation.
    
   \subsection{Kernel weight assignment}\label{sec:kernel_weight}
   
   Unlike rectangular images, omni-directional images have no clear reference direction, so the direction of the kernel becomes ambiguous. To overcome this ambiguity, we set 2 of the 12 vertices of SpherePHD as the north and south poles and use the poles to define the orientation of all pixels to be either upward or downward. We use two kernel shapes which share the same weights for the upward and downward pixels. Furthermore, the order of kernel weight is configured so that the two kernels are 180$^\circ$ rotations of each other. The proposed kernel is expected to learn 180$^\circ$ rotation-invariant properties, making the kernels more robust to rotating environments. We show the feasibility of our kernel design in Sec.~\ref{sec:MNIST_Experiments}, as we compare the performance of our kernel weight assignment, as opposed to other non-rotated kernel assignments.  
   

   \subsection{CNN Implementation}
   \label{subsec:CNN_Implementation}
    
   Utilizing our SpherePHD representation from Sec.~\ref{subsec:Icosa} and our kernels from Sec.~\ref{subsec:SPconv}, we propose a CNN for omni-directional images. We implement our method using conventional CNN implementations that already exist in open deep learning libraries.
   
   Thanks to the SpherePHD representation and our kernels, a convolution layer maintains the image size without padding. Also, a pooled layer has one less subdivision than the previous layer. In other words, the $n^{th}$ subdivision SpherePHD turns into the $(n-1)^{th}$ subdivision SpherePHD after a pooling layer. Depending on the dimension of the desired output, we can adjust the number of subdivision.
  
   \textbf{Convolution layer} 
   To implement the convolution layer in SpherePHD using a conventional 2-dimensional convolution method, we first make the location indices representing the locations of pixels for each subdivision. Using the indices for a given subdivision, we represent the SpherePHD images as 2-D tensors (as opposed to 3-D tensors for conventional images). We then stack the indices of $n$ neighboring pixels in another dimension, and map the corresponding pixel values to the stacked indices (to get an insight about our stacking method, refer to the ``im2col'' function in MATLAB). This makes 3-D tensors where one dimension is filled with $n$ neighboring pixel values of the image. With this representation, we can use conventional 2-D convolution using a kernel of size $1 \times n$. This effectively mimics the convolution on the spherical surface with a kernel of size $n$. The output of the convolution is the 2-D tensors of SpherePHD images with the same subdivision as before. 
   
       \begin{figure}[t]
    \begin{center}
       \includegraphics[width=1.0\linewidth]{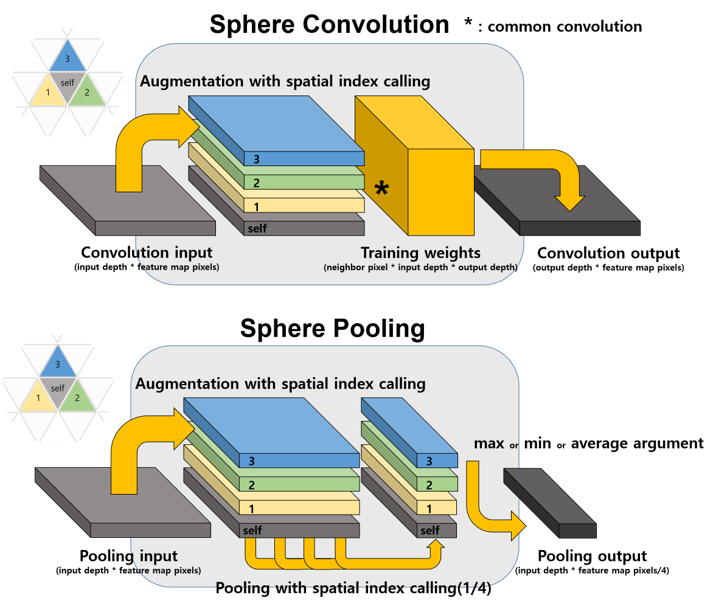}
    \end{center}
    \vspace{-10pt}
       \caption{Tensor-wise implementation of our proposed  convolution and pooling methods.}
    \vspace{-6pt}
    \label{fig:convolution_pooling_explanation}
    \end{figure}

   \textbf{Pooling layer} Implementation of pooling layer is very similar to that of the convolution layer in that we use the same method to stack the neighboring pixels. For images of the $n^{th}$ subdivision, we take the indices of pixels where the pooled values would be positioned in the resulting $(n-1)^{th}$ subdivision images. Then we stack the indices of neighboring pixels to be pooled with, and we map the corresponding pixel values to the indices. After that, the desired pooling operation (\eg{} max, min, average) is applied to the stacked values. The output of pooling is the 2-D tensors of $(n-1)^{th}$ subdivision SpherePHD images.
   
   Figure~\ref{fig:convolution_pooling_explanation} shows a graphical representation of the convolution (upper) and the pooling (lower).

\section{Experiments} \label{sec:Experiments}
     
    We tested the feasibility of our method on three tasks: classification, object detection, and semantic segmentation. We compared the classification performance with MNIST\cite{deng2012mnist} images which are projected onto the SpherePHD, cube map, and ERP representations with random positions and random orientations. We also assessed the object detection and semantic segmentation performances on the SYNTHIA sequence dataset\cite{Ros_2016_CVPR} transformed to aforementioned representations with random tilting. Additionally, we evaluated the semantic segmentation performance on the Stanford2D3D dataset to check the feasibility on real data.

   \subsection{MNIST classification}\label{sec:MNIST_Experiments}
   
    \begin{figure}[t]
    \begin{center}
       \includegraphics[width=0.9\linewidth]{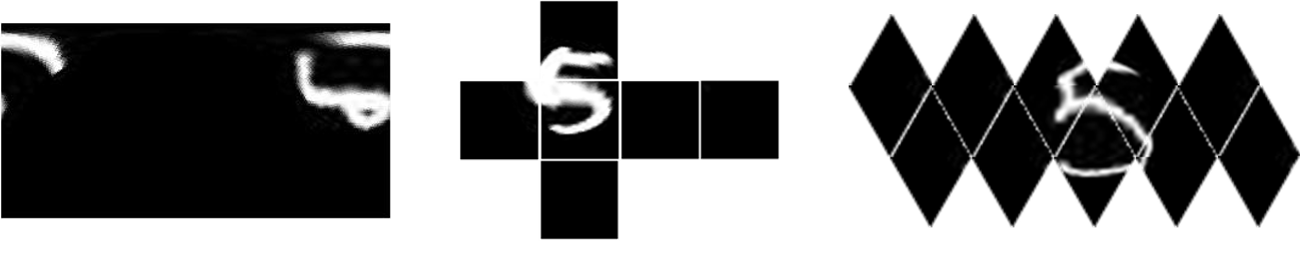}
    \end{center}
    \vspace{-6pt}
       \caption{From left to right: ERP, cube map, SpherePHD representations of MNIST image. These are the three different types of input images. All images represent MNIST digit 5.}
    \label{fig:MNIST_5}
    \vspace{-4pt}
    \end{figure}
    
    \textbf{Dataset} We made three types of spherical MNIST datasets by projecting 70k original MNIST images onto SpherePHD, cube map, and ERP representations with random positions and random orientations as shown in Fig.~\ref{fig:MNIST_5}.
   By randomly changing the position and the orientation of projection, the number of training and test images were augmented from 60k and 10k to 1200k and 700k, respectively.

   We determined the image size of SpherPHD to be 3 times subdivided size, which is  $1 \times 1280(=20\times4^3)$, to maintain similar resolution with $28\times28$ original MNIST images. To maintain the resolution of all datasets consistent, the sizes of cube map and ERP representations were adjusted based on the number of pixels on the equator of the SpherPHD image. The number of pixels along the equator of all datasets became equal by configuring the size of cube map image as $6\times20\times20$ pixels and the size of ERP image as $40\times80$.
    
   \textbf{Implementation details} We designed a single neural network as fully convolutional structure for all three datasets. In other words, we used the same structure but changed the convolution and pooling methods  according to the type of input representation. To compare the performance of the networks with the same structure and the same parameter scale, we used a global average pooling layer instead of using a fully connected layer for the classification task, inspired by  NiN\cite{lin2013network}. Due to the fully convolutional network structure, the parameter scale of the network is only dependent on the kernel size, being independent on the input image size. We also minimized the difference between kernel sizes by using $10\times1$ kernels shown in Fig.~\ref{fig:kernelshape} on our method and $3\times3$ kernels on the other methods. The networks used in the experiment have two pairs of convolution and max pooling layers followed by a pair of convolution and global average pooling layers. 
    
\begin{figure}[t]
    \begin{center}
       \hspace*{-5.1mm}\includegraphics[width=1.1\linewidth]{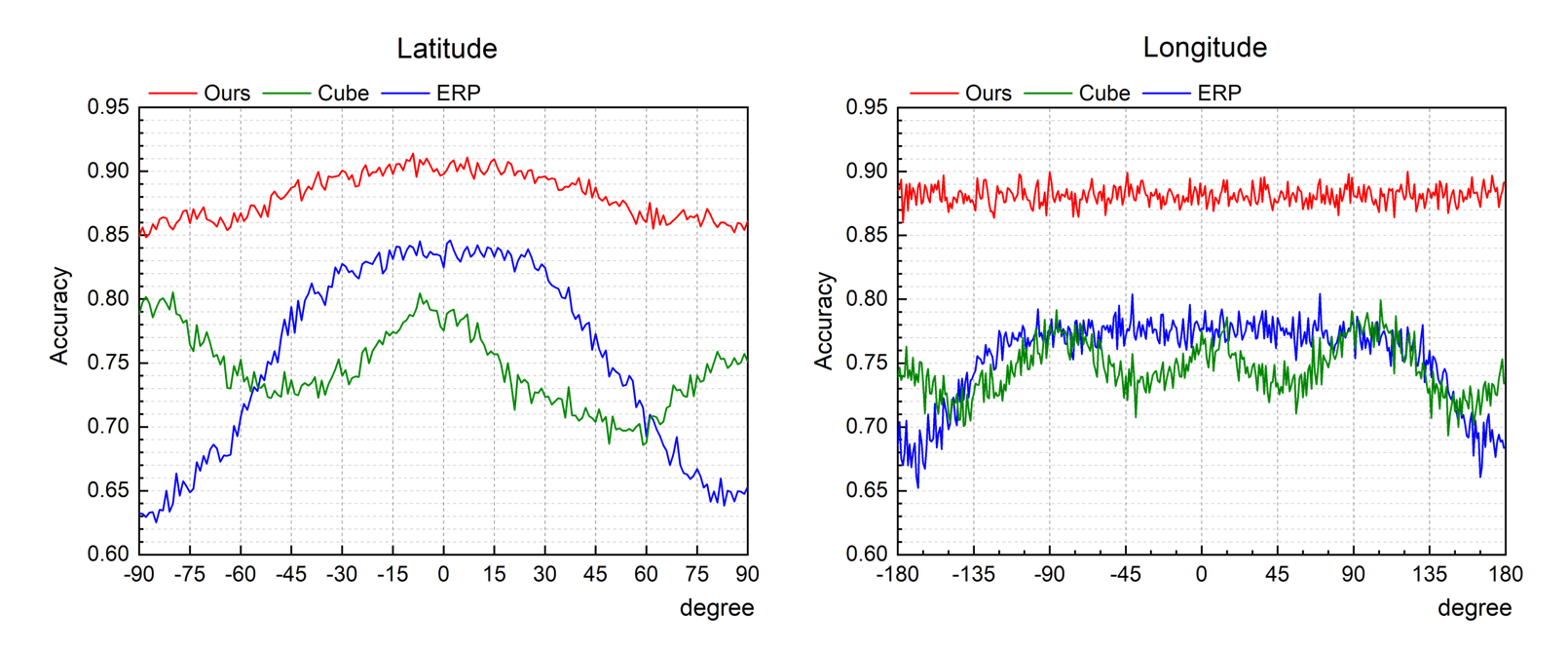}
    \end{center}
    \vspace{-12pt}
       \caption{MNIST classification accuracy for 700k test images along latitude and longitude. The number of samples along latitude and longitude follows uniform distribution. These results are highly related with the  distribution of the irregularity values along latitude and longitude shown in Fig.~\ref{fig:distortion_and_sphere}.}
    \vspace{-2pt}
    \label{fig:MNIST_experiment_phi_theta}
    \end{figure}

\begin{figure}[t]
    \begin{center}
       \includegraphics[width=1\linewidth]{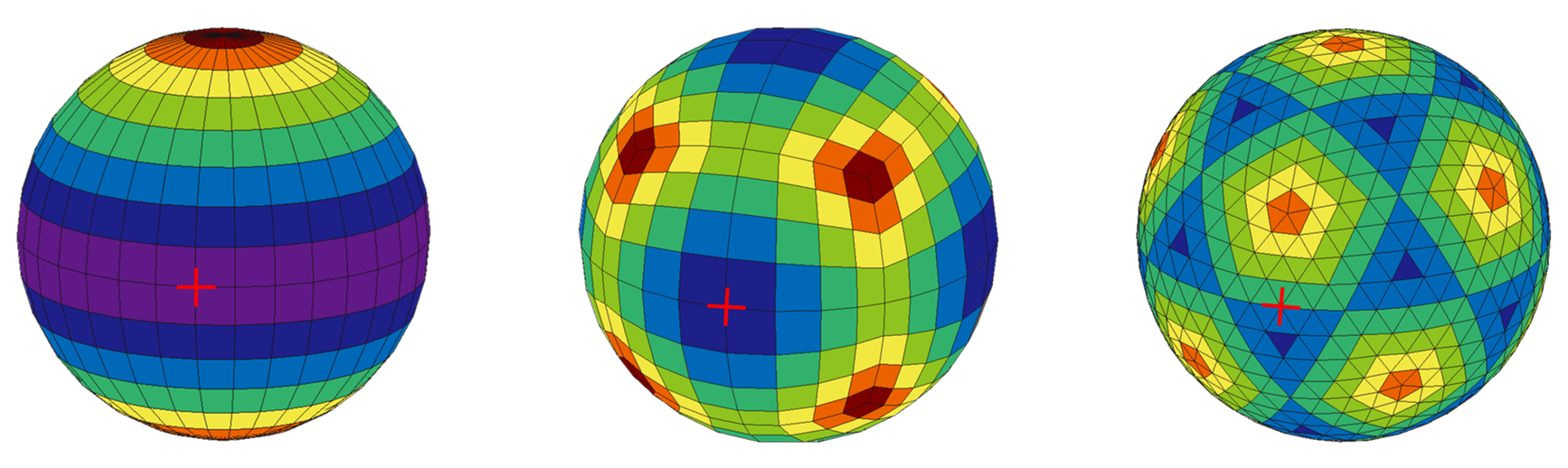}
    \end{center}
    \vspace{-10pt}
       \caption{Distribution of irregularity values defined in Sec.~\ref{subsec:distortion} on the sphere surface. The red cross mark represents the point at ( latitude, longitude)=(0, 0) on the sphere surface. The colors for each representation indicate relative scales of irregularity.}
    \label{fig:distortion_and_sphere}  
    \end{figure}

\begin{figure}[t]
        \begin{center}
        \hspace{0pt}\includegraphics[width=1.0\linewidth]{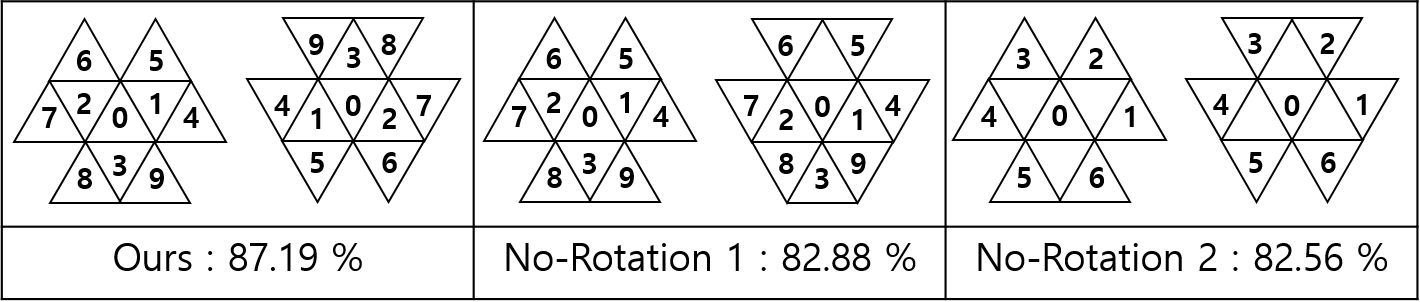}
        \end{center}
        \vspace{-8pt}
        \caption{MNIST classification results with different kernels}
        \label{fig:MNIST_reusults}
        \vspace{-6pt}
    \end{figure}

   \textbf{Result} 
   To understand the effect of irregularity that we defined on each image representation, we measured the performance of the representations according to the positions of MNIST digits in terms of latitude and longitude, as shown in Fig.~\ref{fig:MNIST_experiment_phi_theta}. In the result along the latitude, we can see that our method gives relatively uniform accuracy, although the accuracies near -90\degree{} and 90\degree{} are slightly dropped. This result is consistent with the variation of irregularity of SpherPHD which has relatively small variation of irregularity, as shown in Fig.~\ref{fig:distortion_and_sphere}.  Results for cube map and ERP representations also follow the variation of irregularity. The regions where the variation of irregularity is large (\eg{} edge and vertex) match with the regions where accuracy decreases. In the result along the longitude, we can also see the same tendency but the result from the ERP representation shows additional accuracy drop due to the discontinuity on image boundaries.  Table~\ref{tab:MNIST_Test} shows the overall average accuracies of all representations. In addition, Fig.~\ref{fig:MNIST_reusults} shows the comparison of the performance of our kernel weight assignment to other non-rotated kernel assignments, which shows the effectiveness of our kernel design described in Sec.~\ref{sec:kernel_weight}.
   
    \vspace{-4pt}
    
    \begin{table}[t]
    \centering
    \caption{MNIST classification results of the three methods} \vspace{5pt}
    \scalebox{0.85}{
    \begin{tabular}{|c|c|c|c|}
    \hline 
          & correct predictions & test set size & accuracy (\%) \\ \hline
          SpherePHD & 616,920 & 700,000 & 88.13 \\ \hline
         ERP  & 528,577 & 700,000 & 75.51 \\ \hline
         Cube padding  & 521,937 & 700,000 & 74.56 \\ \hline
    \end{tabular}
    }
    \label{tab:MNIST_Test}
    \end{table}

   \vspace{2pt}
   \subsection{SYNTHIA vehicle detection}\label{sec:synthia_detection}
   
   \textbf{Dataset}
    The SYNTHIA dataset is a vitrual road driving image dataset\cite{Ros_2016_CVPR}. Each sequence consists of front, right, left, and back NFoV images of a driving vehicle, along with the ground truth labels for each object in each scene. Since the camera centers for the four directions are the same at a given moment, we can make a 360\degree{} image for the scene. We projected the scenes onto SpherePHD, cube map, and ERP representations. We conducted two different experiments for the detection task. One is a no-rotation version in which the SYNTHIA images are projected without rotation. The other one is a rotation version created by rotating 360\degree{} SYNTHIA images in random orientations.
    
    \textbf{Implementation details}
    We performed the vehicle detection test using SpherePHD, cube map and ERP representations. To compare the detection accuracies of the representations fairly, we used similar CNN architectures of the same parameter scale (based on YOLO architecture~\cite{Redmon_2016_CVPR}), having the same number of layers. To detect objects in various orientations of 360\degree{} images, we used a bounding circle instead of a bounding box.
    
\begin{figure}[t]
    \centering
    \includegraphics[width=0.9\linewidth]{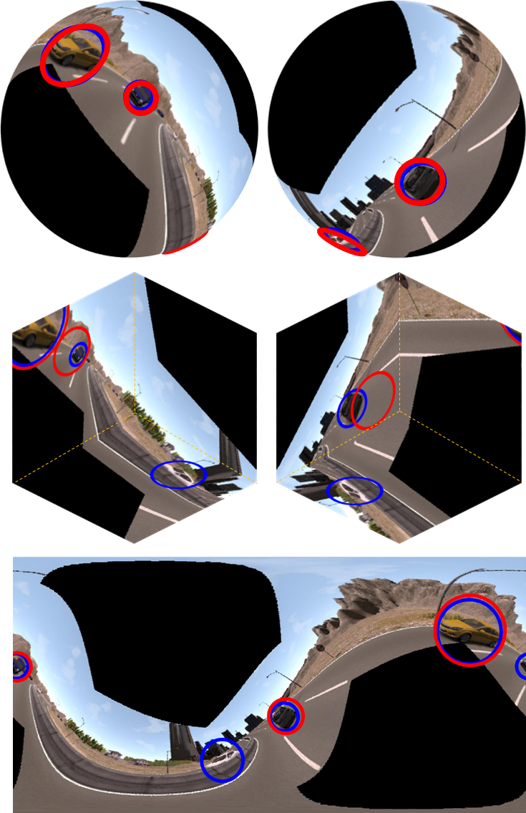}
    \vspace{0pt}
       \caption{Vehicle detection results of the SYNTHIA dataset. From top to bottom, results from SpherePHD, cube map, and ERP representations. Red circles are the predicted bounding circles and blue circles are the ground truths. These images are test samples from the rotation-augmented set.}
    \label{fig:synthia_result}
    \end{figure}

    \textbf{Result}
    As shown in Table~\ref{tab:synthia_result}, when 360\degree{} images (18k train and 4.5k test) are not tilted, where the vehicles are mainly located near the equator, the ERP representation yields higher accuracy than ours. The reason is that, in this case, the shape distortion of target objects near the equator could be negligible. However, when 360\degree{} images (180k, train and 45k test) are tilted, SpherePHD works better while the performance of ERP representation is severely degraded. Even though data augmentation generally increases the performance of network, the detection accuracy of ERP representation decreases on the contrary. Also, the detection accuracy of the cube map representation is lower than ours because the cube map has the discontinuity of kernel orientation at the top and bottom faces of the cube map. Some detection results are shown in Fig.~\ref{fig:synthia_result}. 

    \subsection{Semantic Segmentation}

    \textbf{Dataset}
    We used the Stanford2D3D real-world indoor scene dataset\cite{stanford2D3D} provided in ERP images as well as SYNTHIA driving sequence dataset\cite{Ros_2016_CVPR}. SYNTHIA and Stanford2D3D datasets, which have 16 and 39 classes of semantic labels respectively, are transformed to SpherPHD, cube map, ERP representations. As in Sec.~\ref{sec:synthia_detection}, we also augmented the size of the datasets by tilting the camera randomly.
    
    
    \textbf{Implementation details}
    We designed a neural network as a CNN-based autoencoder structure for all two datasets. We kept the parameter scales of all 360\degree{} image representations consistent. Among many kinds of unpooling methods, We chose a max unpooling method for the decoder. A max unpooling layer of SpherePHD reutilizes the indices used in our pooling layer to get a higher subdivision SpherPHD.     
    
\begin{figure}[t]
    \centering
    \includegraphics[width=1.0\linewidth]{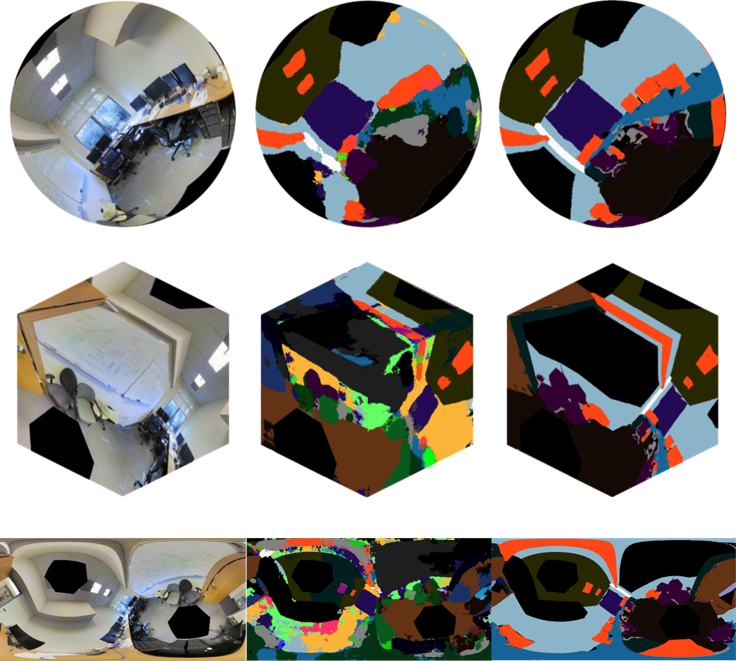}
    \vspace{-10pt}
       \caption{Semantic segmentation results of the Stanford2D3D dataset. From top to bottom: results from SpherePHD, cube map, and ERP representations. From left to right: input image, network output, and ground truth.}
    \label{fig:stanford2D3D_result}
    \end{figure}

    \textbf{Results}
    The evaluation metrics used for semantic segmentation are the average of class accuracies and the overall pixel accuracy. We evaluate on both metrics because the overall pixel accuracy does not necessarily reflect the accuracy of classes that only have few pixels; when classes like the wall and floor, which take up a large portion of the pixels, have high accuracies the overall accuracy can be skewed. Thus we also evaluated on the average of class accuracies.
    Table~\ref{tab:stanford2D3D_result} shows the quantitative results of the semantic segmentation experiment. For both datasets, our SpherPHD method outperforms other methods. Even if the accuracies from the Stanford2D3D dataset are much lower than the accuracies from SYNTHIA due to the much higher number of classes and the noise in real-world data, our method still maintains higher accuracies than other methods with large gaps.  
    Figure~\ref{fig:stanford2D3D_result} shows the semantic segmentation results for different representations. 

    \begin{table}[t]
    \caption{Detection average precision (AP) of three different image representations (\%)}
    \vspace{4pt}
    \centering
    \begin{adjustbox}{width=0.83\columnwidth,center}
\begin{tabular}{|c|c|c|c|}
\hline                  
& {SpherePHD} & {ERP}   & {cube map} \\ \hline
{\begin{tabular}[c]{@{}c@{}}SYNTHIA\end{tabular}} & {43.00}     & {56.04} & {30.13}    \\ \hline
{\begin{tabular}[c]{@{}c@{}}SYNTHIA\\ (rotation-augmented)\end{tabular}} & {64.52}     & {39.87} & {26.03}    \\ \hline
\end{tabular}
    \end{adjustbox}
    \label{tab:synthia_result}
    \end{table}

    \begin{table}[t]
    \vspace{-3pt}
    \caption{The average of class accuracies and the overall pixel accuracy of three different image representations (\%)}
    \vspace{2pt}
    \begin{adjustbox}{width=\columnwidth,center}
    \setlength{\tabcolsep}{0.3em}
    \begin{tabular}{|c|c|c|c|c|c|c|}
\hline
& \multicolumn{2}{c|}{SpherePHD} & \multicolumn{2}{c|}{ERP}     & \multicolumn{2}{c|}{cube map} \\ \hline
 & per class  & overall  & per class & overall & per class & overall \\ \hline
SYNTHIA                                                             & 70.08      & 97.20    & 62.69     & 95.07   & 36.07     & 66.04   \\ \hline
\begin{tabular}[c]{@{}c@{}}Stanford2D3D\\ (real dataset)\end{tabular} & 26.40      & 51.40    & 17.97     & 35.02   & 17.42     & 32.38   \\ \hline
\end{tabular}
    \end{adjustbox}
    \label{tab:stanford2D3D_result}
    \vspace{-2pt}
    \end{table}

\section{Conclusion} \label{sec:Conclusion}
    Despite the advantages of 360$^\circ$ images, CNNs have not been successfully applied to 360$^\circ$ images because of the shape distortion due to nonuniform resolving power and the discontinuity at image borders of the different representation methods. To resolve these problems, we proposed a new representation for 360$^\circ$ images, SpherePHD. The proposed representation is based on a spherical polyhedron derived from an icosahedron, and it has less irregularity than ERP and cube map representations. We also proposed our own convolution and pooling methods to apply CNNs on the SpherePHD representation and provided the details of these implementations, which allow to apply the SpherePHD representation to existing CNN-based networks. 
    Finally, we demonstrated the feasibility of the proposed methods through classification, detection, and semantic segmentation tasks using the MNIST, SYNTHIA, and Stanford2D3D datasets.

\section*{Acknowledgement} 
This work was supported by Samsung Research Funding Center of Samsung Electronics under Project Number SRFC-TC1603-05 and National Research Foundation of Korea (NRF) grant funded by the Korea government (MSIT) (NRF-2018R1A2B3008640).

\small
\bibliographystyle{ieee}
\bibliography{egbib}

\end{document}